\documentclass{article}
\usepackage{graphicx} 
\usepackage{twocolceurws}
\usepackage{amsmath}
\usepackage{algorithm}
\usepackage{algpseudocode}
\usepackage{makecell} 
\usepackage{color}
\usepackage{amssymb}
\usepackage[font=small,labelfont=bf]{caption}
\usepackage{multirow}
\usepackage{verbatimbox}
\usepackage{appendix}
\usepackage{url}

\title{Ctrl-A: Control-Driven Machine Learning using Adaptive Online Data Augmentation}
\author{Jesper B.~Christensen$^1$, Ciaran Bench$^2$, Spencer A.~Thomas$^2$, Hüsnü Aslan$^1$, \\ David Balslev-Harder$^1$,  Nadia A.~S.~Smith$^{3,4}$, and Alessandra Manzin$^5$ }
\institution{$^1$Danish Fundamental Metrology (DFM), Kogle Allé 5, DK-2970 Hørsholm, Denmark \\
$^2$National Physical Laboratory (NPL), Teddington,  United Kingdom  \\
$^3$TÜV SÜD, Warrington, United Kingdom \\
$^4$Royal Surrey NHS Foundation Trust, Guildford, United Kingdom\\
$^5$Istituto Nazionale di Ricerca Metrologica (INRiM), Strada delle Cacce, 91, 10135 Torino, Italy}
\date{\today}

\begin{document}
\maketitle
\begin{abstract}
    We introduce ControlAugment (Ctrl-A), an automated data augmentation algorithm for image-vision tasks, which incorporates principles from control theory for online adjustment of augmentation strength distributions during model training. Ctrl-A eliminates the need for initialization of individual augmentation strengths. Instead, augmentation strength distributions are dynamically, and individually, adapted during training based on a control-loop architecture and what we define as relative operation response curves. Using an operation-dependent update procedure provides Ctrl-A with the potential to reduce the use of augmentation operations that negatively impact model performance, alleviating the need to manually engineer augmentation policies for new image classification tasks. Experiments on the CIFAR-10, CIFAR-100, SVHN-core, and Tiny ImageNet benchmark datasets using WideResNet-28-10 and ResNet architectures demonstrate that Ctrl-A provides competitive performance across a range of image classification benchmarks.
\end{abstract}

\section{Introduction}
Computer vision (CV) has advanced rapidly with the evolution of artificial intelligence (AI) \cite{zhao2024review}. AI-powered solutions based on deep neural network (DNN) architectures are now widely applied to tasks such as medical image processing (e.g., radiology, histopathology, and radiotherapy planning) \cite{panayides2020ai,rajpurkar2022ai}, biometric facial recognition \cite{minaee2023biometrics}, and quality-assurance inspection in Industry 4.0 \cite{peres2020industrial}. DNN models typically consist of tens or hundreds of millions of parameters to effectively accomplish complex image processing and analysis tasks. However, their effectiveness depends on access to large-scale training datasets, as insufficient data can lead to overfitting, where the model learns dataset-specific patterns and fails to generalize to unseen inputs.

Data augmentation (DA) has proven to be an effective approach for mitigating overfitting and improving the generalization performance of DNN models in CV tasks \cite{shorten2019survey,mumuni2022data,xu2023comprehensive, goodfellow2014explaining,du2024joypose,wang2025data}. DA employs various geometric and color-based transformations, as well as adversarial perturbations \cite{goodfellow2014explaining, miyato2018virtual,ntelemis2023generic}, to increase training-data diversity and improve model generalization. Geometric and color-based transformations, such as rotations, translations, brightness, and contrast adjustments, are global transformations that modify the image as a whole and are widely used to enrich the training data while assuming label preservation \cite{kumar2024image}. Their effectiveness, however, depends on selecting appropriate operations and related augmentation strengths for the task at hand.

Despite their effectiveness, augmentation operations and strengths are often selected based on prior studies or established augmentation policies, rather than being systematically tailored to the data and learning problem \cite{shorten2019survey,mumuni2022data,xu2023comprehensive}. Determining suitable transformations and their strengths manually can require substantial domain knowledge and time-consuming trial-and-error experimentation \cite{cubuk2019autoaugment}.

These challenges have motivated the development of automated DA strategies, including AutoAugment (AA) \cite{cubuk2019autoaugment}, which demonstrated the possibility of learning augmentation sub-policies from a proxy AI task. However, many automated DA approaches introduce significant computational overhead through the search or optimization required to determine augmentation policies subsequently used for training the target model. An alternative is to adapt augmentation policies online during model training, using information from the evolving model to update augmentation operations and their strengths without a separate policy-search phase.

In this work, we formulate online adaptive DA as a feedback-control problem. Our algorithm, named \textit{ControlAugment} (Ctrl-A), utilizes the regularizing effect of DA to individually update augmentation strengths for each augmentation operation as model training progresses through successive phases. Ctrl-A incurs limited computational overhead, with computational cost close to that of TrivialAugment \cite{muller2021trivialaugment}, and requires no specific initialization of the augmentation strengths.

In particular, this paper makes the following contributions to the field of automatic DA:
\begin{itemize}
    \item Drawing on concepts from control theory, we formulate an online DA strategy that serves as an adaptive model regularizer by dynamically controlling the ratio between training and validation loss during learning.    
    \item We introduce parametrized augmentation-strength distributions (ASDs) and propose an online update procedure for their parameters, guided by a newly defined concept termed \textit{relative operation response}, which quantifies a model's response to individual operations.
    \item We show that a commonly used WideResNet-28-10 training setup on CIFAR-10 can mask differences between DA methods. Through controlled experiments, we demonstrate that training setup choices can substantially influence the observed relative performance of augmentation methods, highlighting the importance of carefully designed benchmark protocols.
\end{itemize}
    


\section{Related Work}
Automated DA methods aim to reduce the need for manually designing augmentation policies by automatically selecting augmentation operations and their strengths. In this section, we review representative approaches to automated DA and distinguish between methods that determine augmentation policies through a preceding search or optimization procedure and methods that adapt augmentation policies online during model training.

\subsection{Offline methods}
We first consider automated DA approaches in which augmentation policies are determined through a dedicated search or optimization process, rather than being continuously adapted through feedback within the training process of the target model.

One of the most influential methods in automated DA is AutoAugment (AA) \cite{cubuk2019autoaugment}, which formulates augmentation policy search as a discrete optimization problem, using reinforcement learning to discover effective transformation strategies. While novel, the computational overhead in finding paired augmentations (sub-policies) scales poorly due to the exponentially growing search space, defined by the number of transformations and augmentation strength levels. 

To reduce the computational overhead in learning relevant DA policies, several follow-up methods were developed, including Fast AutoAugment (Fast-AA) \cite{lim2019fast}, Faster AutoAugment (Faster-AA) \cite{hataya2020faster}, Deep AutoAugment \cite{zheng2022deep}, and Population-based Augmentation (PBA) \cite{ho2019population}. Both Fast- and Faster-AA  avoid the use of reinforcement learning and replace it with more efficient methods. Fast-AA employs density matching and attempts to find augmentation policies that make the training data resemble a held-out validation set \cite{lim2019fast}. Faster-AA, on the other hand, shares similarities with other differentiable methods \cite{li2020differentiable,liu2021direct} by formulating augmentation search as a continuous gradient-based optimization problem \cite{hataya2020faster}. PBA takes a different approach as it trains a population of models in parallel, allowing them to investigate different augmentation policies, and incorporates model-parameter sharing at specific training intervals to transfer beneficial augmentation strategies \cite{ho2019population}. This enables PBA to apply different augmentation policies at different training stages, as one global policy is unlikely to be optimal. Like Fast-AA and Faster-AA, PBA improves the efficiency of automated DA learning while retaining the bi-operation sub-policy structure introduced in AA \cite{cubuk2019autoaugment}.

In contrast to the algorithms described above, which rely on searching for augmentation policies composed of predefined operation pairs and magnitudes, RandAugment (RA) \cite{cubuk2020randaugment} and TrivialAugment (TA) \cite{muller2021trivialaugment} introduce a simpler approach to automated DA. Rather than solving an optimization problem or a proxy reinforcement learning task to determine efficient DA sub-policies, RA reduces the task to a simple two-dimensional search over the number of operations $N$ and their (common) strength $M$ \cite{cubuk2020randaugment}. With this approach, $N$ operations are randomly selected for each training sample and applied with augmentation strength $M$, thereby emphasizing training data diversity. Rather than relying on curated augmentation sub-policies, RA therefore uses random operation selection and a common augmentation strength to obtain effective data augmentation with reduced policy search. TA further simplifies this approach by removing the policy search phase entirely. Instead, it works by blindly selecting a single augmentation operation and randomly sampling its strength for each training sample \cite{muller2021trivialaugment}. Despite this simplification, TA has been shown to achieve performance comparable to RA in several benchmark settings \cite{muller2021trivialaugment}.

\subsection{Online adaptive methods}
Recent automated DA research has increasingly explored online approaches, in which augmentation policies are adapted during model training rather than determined \textit{a priori}. Several approaches have been investigated, including Meta AutoAugment (Meta-AA) \cite{zhou2021metaaugment}, Online Hyperparameter Learning (OHL) \cite{lin2019online}, Universal Adaptive Data Augmentation (UADA) \cite{xu2022universal}, OnlineAugment \cite{tang2020onlineaugment}, Diversity-based data Augmentation (DivAug) \cite{liu2021divaug}, and Invariance-Constrained Learning (ICL) \cite{hounie2023automatic}. The adaptation mechanisms used, however, vary considerably: Meta-AA jointly optimizes model parameters and DA policies using implicit gradient-based optimization \cite{zhou2021metaaugment}; OHL formulates DA learning as a hyperparameter-learning task \cite{lin2019online}; UADA uses an adversarial approach to update DA parameters by maximizing the training loss \cite{xu2022universal}; OnlineAugment employs augmentation models that learn efficient augmentations while being trained in alternation with the target model \cite{tang2020onlineaugment}; \mbox{DivAug} formulates a metric ``variance diversity'' which, based on the model state, is used to enhance augmentation diversity in the training data \cite{liu2021divaug}; and ICL formulates DA as an invariance-constrained learning problem solved using Markov-chain Monte Carlo sampling and avoids undesired augmentations \cite{hounie2023automatic}.

Related online approaches have explored dynamically adjusting augmentation magnitudes or operation probabilities during training. For example, EntAugment dynamically adjusts augmentations based on an auxiliary entropy loss term \cite{yang2024entaugment}, while AdaAugment introduces a dual-model architecture to jointly optimize a policy network for online adaptation of augmentation magnitudes \cite{yang2025adaaugment}. Finally, a recent paper on cervical cytopathology makes active use of augmented validation data to optimize the probability distribution over the augmentation space based on validation performance \cite{TANG202593}. Our method, Ctrl-A, similarly uses augmented validation data, but explicitly characterizes the model's response to individual augmentation operations and uses the resulting response curves to achieve online adaption of operation-specific augmentation-strength distributions through a closed-loop feedback-control mechanism.


\section{ControlAugment}
\subsection{Setting}
We consider the standard supervised image classification task, in which each image $x \in \mathcal{X}$ in the dataset $\mathcal{X} \subset \mathbb{R}^{n\times m \times c} $ is associated with a single class label $y \in \mathcal{Y} =  \lbrace 1, ..., C \rbrace$, with $c$ denoting number of (color) input channels and $C$ being the number of output labels. In the simplest possible case, the paired dataset is divided into a training set $\mathcal{D}_\mathrm{Train} = (x_\mathrm{Train}, y_\mathrm{Train}  )$ and a test set $ \mathcal{D}_{\mathrm{Test}} = (x_{\mathrm{Test}}, y_{\mathrm{Test}}  )$. In this case, our goal is to learn model parameters $\mathbf{\theta} $ for a chosen model $f_\theta : \mathcal{X} \rightarrow \mathcal{Y}$, which generalizes such that it approximately minimizes the expectation value of a loss function, $\mathcal{L}(f_\theta (x_{\mathrm{Test}}), y_{\mathrm{Test}})$, while only allowing the model to be explicitly trained on dataset $\mathcal{D}_{\mathrm{Train}}$. 

In addition to the training and test datasets, a validation dataset, $\mathcal{D}_\mathrm{Val} = (x_\mathrm{Val}, y_\mathrm{Val}  )$, is commonly used in practice during training to gauge model performance on data that the model is not explicitly trained on. As we describe shortly, the validation dataset plays an active role in the Ctrl-A framework. 

\subsection{Concept and framework}

\begin{figure}[t]
    \centering
    \includegraphics[width=0.98\linewidth]{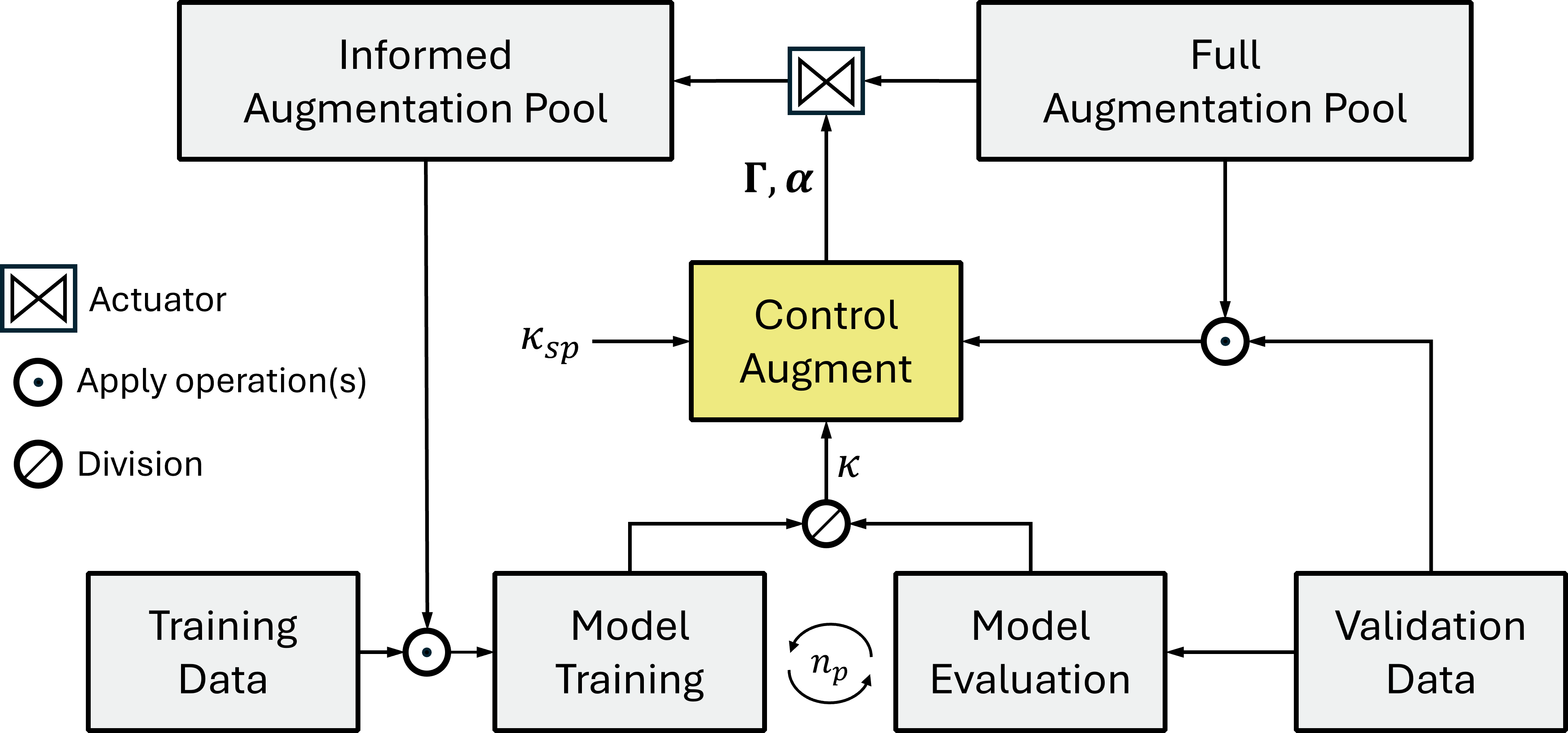}
    \caption{\label{fig:IAsketch}Illustration of the ControlAugment framework. Model training proceeds in phases of $n_p$ epochs, separated by the ControlAugment algorithm, which regulates the informed augmentation pool by adjusting the augmentation strength parameters $\boldsymbol{\Gamma}$ and $\boldsymbol{\alpha}$ for the subsequent training phase. The adjustment is executed by the ControlAugment block which quantifies the response of each operation using augmented validation data and through an internal control parameter ($\xi$) which is updated by comparing the setpoint value ($\kappa_{sp}$) with a relative training/validation performance metric ($\kappa$). 
    }
\end{figure}

Our automatic DA-based training procedure for supervised learning is illustrated in Fig.~\ref{fig:IAsketch}. A central concept in Ctrl-A is the distinction between a \textit{full augmentation pool} and an \textit{informed augmentation pool}, both comprising the same $K$ operation types. The full augmentation pool defines the maximum strength each operation can attain in principle, whereas the periodically updated informed pool specifies the subset of augmentation strengths available for generating augmented training samples. Depending on the augmentation space considered, the full augmentation pool, which remains fixed throughout training, may correspond to the standard augmentation space introduced for RA \cite{cubuk2020randaugment}, the wide augmentation space used for TA \cite{muller2021trivialaugment}, or the Control augmentation space introduced in this work, comprising the operations listed and parametrized in Appendix~A, Table~\ref{tab:IAaugpool}.

In Ctrl-A, each operation $\mathcal{O}_i \in \mathcal{O}$ is parametrized with a normalized augmentation strength $\gamma_i \in [0,1]$, for $i \in \lbrace1,2,...,K\rbrace$. Moreover, an operation, applied to image $x$ as $\mathcal{O}_i(x;\gamma_i)$, is characterized by (\textit{1}) yielding a progressively stronger perturbation to image $x$ as $\gamma_i$ increases from zero to unity, and (\textit{2}) simplifying to the identity operator in the limit of vanishing augmentation strength, i.e.,~${\lim_{\gamma_i \to 0} \mathcal{O}_i(x; \gamma_i) = x}$. 

During model training using Ctrl-A, every single training sample is an augmented version, $x_A$, of original training data, $x$. A single image augmentation, $ x_{\mathcal{A}}$, is obtained by (\textit{1}) randomly selecting $N$ of the $K$ operations (sampling without replacement) with integer indices $i_1, i_2, ..., i_N$ from $\mathcal{O}$, (\textit{2}) randomly drawing for each index, $i_n$, the operations' augmentation strength $\gamma_{i_n}  \sim U_{\alpha_{i_n}}(0,\Gamma_{i_n})$, and finally (\textit{3}) applying composite operations
\begin{equation}
    x_\mathcal{A} = \mathcal{O}_{i_N} ( ... (\mathcal{O}_{i_2}(\mathcal{O}_{i_1}(x;\gamma_{i_1});\gamma_{i_2})...);\gamma_{i_N}).
\end{equation}
We refer to the case of applying $N$ operations as \mbox{CtrlA($N$)}, and note that different operations, in general, do not commute, which further enhances training-data diversity for $N>1$. 

\begin{figure}[th]
    \centering
    \includegraphics[width=1\linewidth]{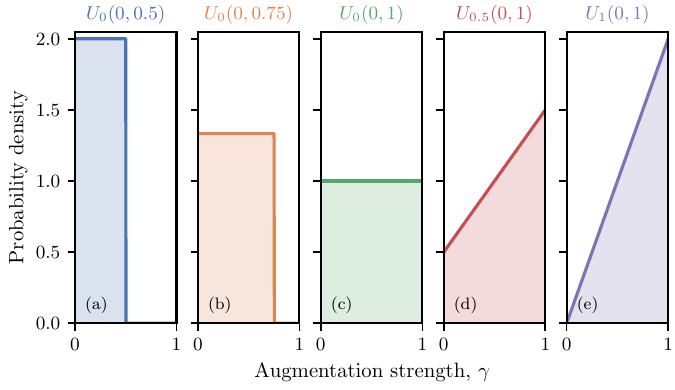}
    \caption{Examples of augmentation strength distributions with increasing distribution means from left to right.}
    \label{fig:ASDs}
\end{figure}
The probability distribution, $U_\alpha(0,\Gamma)$, from which individual augmentation strengths are drawn, plays an important role in Ctrl-A, and we refer to it as the augmentation strength distribution (ASD). It is here parametrized by two values, the distribution upper bound, $\Gamma \in[0,1]$, and the skewness value, $\alpha \in [0,1]$. For $\alpha = 0$, $U_0(0,\Gamma) = U(0,\Gamma)$ is the uniform distribution with continuous support $[0,\Gamma]$, and for $\alpha=1$, $U_1(0,\Gamma)$ is the triangular distribution with mode $\Gamma$. Examples of $U_\alpha(0,\Gamma)$ are visualized in Fig.~\ref{fig:ASDs}, demonstrating the capability of the ASD to shift weight toward larger augmentation strengths as $\Gamma$ and $\alpha$ are increased. The distribution mean is given by $ (1+\alpha/3)\Gamma/2$ and is thereby a growing function of both $\Gamma$ and $\alpha$. As a result, the ensemble strength of DA, for each operation $\mathcal{O}_i$, can be controlled through the individual values for $\Gamma_i$ and $\alpha_i$. Notably, in the special case of $\boldsymbol{\Gamma} = \boldsymbol{1}$, $\boldsymbol{\alpha} = \bold{0}$, and $N=1$, we retrieve the constant setting applied in TrivialAugment \cite{muller2021trivialaugment}.

Finally, Ctrl-A partitions model training into training phases, indexed $j$, with each phase lasting $n_p$ epochs. A phase is characterized by constant vectors $\boldsymbol{\Gamma} = (\Gamma_1, \Gamma_2, ...,\Gamma_K)$ and $\boldsymbol{\alpha}=(\alpha_1,\alpha_2,...,\alpha_K)$, and is concluded by re-evaluating $\boldsymbol{\Gamma}$ and $\boldsymbol{\alpha}$ ahead of the following training phase. This re-evaluation enables the degree of DA to be dynamically updated during training, and is discussed next.

\subsection{Updating the ASDs }
\label{IA_start}

In succession to every training phase, we employ a procedure for computing $\boldsymbol{\Gamma}^{(j+1)}$ and $\boldsymbol{\alpha}^{(j+1)}$, resulting in an updated set of ASDs, i.e.~$U_{\alpha_i}(0,\Gamma_i)$ for $i \in \lbrace1,2,...,K\rbrace$. The determination of $\boldsymbol{\Gamma}^{(j+1)}$ and $\boldsymbol{\alpha}^{(j+1)}$ relies on the introduction of the relative operation response (ROR), $R_{\mathcal{O}_i}(\gamma)$ defined as
\begin{equation}
    R_{\mathcal{O}_i}(\gamma_i) = \frac{\mathrm{Acc}_f(\mathcal{O}_i (\mathcal{D}_\mathrm{Val} ; \gamma_i))}{\mathrm{Acc}_f(\mathcal{D}_\mathrm{Val})} ,
\end{equation}
in which the numerator, $\mathrm{Acc}_f(\mathcal{O}_i (\mathcal{D}_\mathrm{Val} ; \gamma_i))$, is the augmented classification accuracy, obtained by computing the validation accuracy after the operation $\mathcal{O}_i( \,\cdot \, ; \gamma_i)$ has been applied, with strength $\gamma$, to the entire dataset, whereas the denominator is the classification accuracy on the unaugmented validation set. Due to the parametrization of all operations, the ROR immediately satisfies $R_{\mathcal{O}_i}(0) = 1$ for all $\mathcal{O}_i \in \mathcal{O}$, and its deviation from unity with increasing $\gamma$ quantifies the effect of each operation type.

\subsubsection{Determining $\boldsymbol{\Gamma}^{(j+1)}$}
The ASD mean depends on $\Gamma$ to first order, and so $\Gamma$ is also considered the primary ASD parameter. For each $\mathcal{O}_i \in \mathcal{O}$, the upper bound parameter $\Gamma_i^{(j+1)}$ is set to the value $\gamma_i$ that solves the implicit equation
\begin{equation}
     R_{\mathcal{O}_i}(\gamma_i)   - \xi ^{(j+1)} = 0\,   ,
    \label{eq:IArule}
\end{equation}
where $\xi \in [0,1]$ is a control parameter (see Sec.\,\ref{sec:xi}). Equation~\ref{eq:IArule} has a simple graphical explanation as demonstrated in Fig.~\ref{fig:IArule}, and its solution is estimated, using regression analysis, after evaluating the ROR for $\gamma_i=\{0,\Delta \gamma, 2\Delta \gamma, \dots, 1\}$. With this formulation, the available augmentation strengths in the next training phase are individually tailored depending on how sensitive the model and data is to each individual operation. 
Cases in which $\min(R_{\mathcal{O}_i}(\gamma_i)) > \xi^{(j+1)}$ are dealt with by setting ${\Gamma_i^{(j+1)}=1 }$. Further details regarding the practical implementation are provided in Appendix B. 

\subsubsection{Determining $\boldsymbol{\alpha}^{(j+1)}$}
\begin{figure}
    \centering
    \includegraphics[width=1\linewidth]{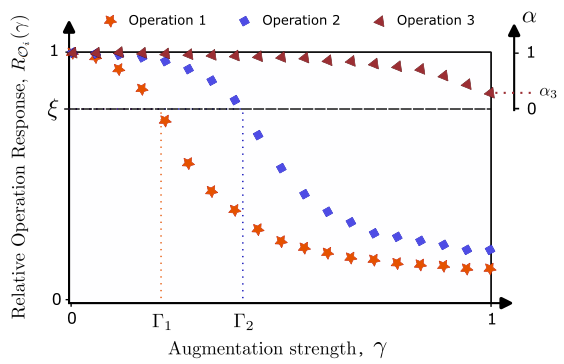}
    \caption{Illustration of the Ctrl-A update procedure (Eqs.~\ref{eq:IArule} and \ref{eq:2ndIArule}), which relies on relative operation response curves to determine new values for the ASD parameters $\Gamma$ and $\alpha$ based on the value of the control parameter, $\xi$. Implicitly, $\Gamma_3 = 1$ and $\alpha_1=\alpha_2 = 0$.}
    \label{fig:IArule} 
\end{figure}

The ASD mean does not have a leading-order dependence on $\alpha$, so $\alpha$ can also be referred to as the secondary ASD parameter. Similarly to $\Gamma_i^{(j+1)}$, the determination of $\alpha_i^{(j+1)}$ depends on the control parameter $\xi^{(j+1)}$ and is computed according to the heuristic rule

\begin{equation}
    \alpha_i^{(j+1)} = \frac{ R_{\mathcal{O}_i} (\gamma_i=1) - \xi^{(j+1)}}{1-\xi^{(j+1)}} ~,
    \label{eq:2ndIArule}
\end{equation}
and clipped to the unit interval $[0,1]$ when necessary.
With this definition, which is also visualized in Fig.~\ref{fig:IArule}, we choose to keep $\alpha_i^{(j+1)} = 0$ unless $\Gamma_i^{(j+1)}  = 1$, i.e.~adding a skew to the ASD only if $R_{\mathcal{O}_i}(1) > \xi^{(j+1)} $. The maximum skew of $\alpha_i = 1$ is obtained if the model performance is unaffected (or improved) by operation $\mathcal{O}_i$, i.e.~if $R_{\mathcal{O}_i}(1) \geq R_{\mathcal{O}_i}(0) = 1$

\subsection{Parameter control algorithm}
\label{sec:xi}
The parameter $\xi^{(j)}$ controls the strength of DA in phase $j$ through Eqs.~\ref{eq:IArule} and \ref{eq:2ndIArule}. Smaller values of $\xi$ correspond to stronger augmentations, while values close to unity result in weak augmentations. The optimal region of $\xi$ is expected to depend both on task, model architecture, number of operations, training state, and dataset representativeness and size, motivating adaptive adjustment during training.

To enable online adaptation, Ctrl-A relies on two well-established effects of data augmentation: (\textit{1})~its regularizing effect, which mitigates overfitting to the training data \cite{liu2021divaug}, and (\textit{2}) the degradation that can arise from excessively strong transformations, which may homogenize samples across classes, leading to implicit class mixing and impaired discriminative learning \cite{shorten2019survey}. To balance these two effects, we formulate the control objective
\begin{equation}
     \min_{\xi} ~\left|   \hat{\mathcal{L}}_{f}\left( \mathcal{D}_\mathrm{Train}^{\mathrm{aug}}  \right) - \kappa_{sp}  \hat{\mathcal{L}}_f (\mathcal{D}_\mathrm{Val} )   \right| ,
\label{eq:minimization}
\end{equation}
where $\mathcal{L}$ is a non-negative convex loss function (e.g.~cross entropy), the hat notation, $\hat{\scriptstyle{\bullet}} $, denotes the averaged loss across a control phase, and $\kappa_{sp}$ defines the target control setpoint that determines the desired imbalance between training and validation loss. Depending on the choice of $\kappa_{sp}$, Ctrl-A can be operated in three regimes: strong augmentation ($\kappa_{sp} > 1$), balanced augmentation ($\kappa_{sp} \approx 1$), and weak augmentation ($\kappa_{sp} < 1$).

The proposed control strategy exploits the observation that the training loss can be influenced through the general strength of data augmentation. To address the control objective in Eq.~\ref{eq:minimization}, we define the measured signal (process variable) as
\begin{equation}
\kappa^{(j)} =  \frac{ \hat{\mathcal{L}}_f\left( \mathcal{D}_{\mathrm{Train}}^{\mathrm{aug}}  \right)}{\hat{\mathcal{L}}_f (\mathcal{D}_{\mathrm{Val}} )},
\label{eq:kappa}
\end{equation}
which feeds the controller update according to
\begin{equation}
\xi^{(j+1)} = \xi^{(j)} + \Delta \xi = \xi^{(j)} +K_I ^{(j)}\left(  \kappa^{(j)} - \kappa_{sp} \right).
\label{eq:IAhyperparameter}
\end{equation}
Here, $K_I^{(j)}$ denotes the integral gain of phase $j$. Rather than using a constant gain, Ctrl-A employs a state-dependent integral gain schedule of the form
\begin{equation}
K_I^{(j)} = k_I (1-\xi^{(j)}),
\end{equation}
where $k_I > 0$ is a user-defined gain coefficient. This formulation reduces the controller gain as $\xi$ approaches unity, while allowing progressively larger updates for decreasing values of $\xi$. 

From Eq.~\ref{eq:IAhyperparameter}, the controller exhibits the following behavior: when the phase-averaged training loss exceeds the scaled validation loss ($\kappa^{(j)} > \kappa_{sp}$), the control increment $\Delta \xi$ becomes positive, leading to weaker augmentations. Conversely, when $\kappa^{(j)} < \kappa_{sp}$, $\Delta \xi$ is negative, resulting in stronger augmentations. After updating $\xi$, the ASD parameters $\boldsymbol{\Gamma}$ and $\boldsymbol{\alpha}$ are recomputed for the next training phase.

\begin{algorithm}[h]
\caption{Ctrl-A Pipeline}
\label{alg:ctrla_dense}
\begin{algorithmic}
\State \textbf{Input:} $\mathcal{D}_{\mathrm{Train}}$, $\mathcal{D}_{\mathrm{Val}}$, Space $\mathcal{O}$, Number of augmentations $N$, Target gap $\kappa_{sp}$, Phase length $n_p$
\State \textbf{Initialize:} State $\xi^{(0)}$, ASD parameters $\boldsymbol{\Gamma}^{(0)}, \boldsymbol{\alpha}^{(0)}$
\For{phase $j = 0, 1, 2, \dots$} 
    \State \textbf{// Step (i): Model Training}
    \State Train for $n_p$ epochs on $\mathcal{D}_{\mathrm{Train}}$ using $(\boldsymbol{\Gamma}^{(j)},\boldsymbol{\alpha}^{(j)})$

    \State Evaluate average train/validation loss ratio $\kappa^{(j)}$
    
    \State \textbf{// Step (ii): Controller Update}
    \State Update controller state $\xi^{(j+1)}$ using Eq.~\ref{eq:IAhyperparameter}
    
    \State \textbf{// Step (iii): ASD Adaptation}
    \For{each operation $\mathcal{O}_i$ ($i=1 \dots K$)}
        \State Compute ROR curve via inference on $\mathcal{D}_{\mathrm{Val}}$
        \State Update $\Gamma_i^{(j+1)},\alpha_i^{(j+1)}$ using Eqs.~\ref{eq:IArule}, \ref{eq:2ndIArule} 
    \EndFor
\EndFor
\end{algorithmic}
\end{algorithm}

Fundamentally, Ctrl-A operates as a closed-loop control system that dynamically adjusts per-operation DA strength to regulate the relationship between training and validation loss. The detailed procedural implementation of this closed-loop mechanism, which differentiates Ctrl-A from other methods such as e.g.~RA and TA, is formalized in Algorithm~\ref{alg:ctrla_dense}.

\subsection{An extended role for the validation dataset}
In standard ML settings, the validation dataset, $\mathcal{D}_\mathrm{Val}$, serves a passive role by monitoring the generalization performance of the model $f_\theta$ during training and supporting hyperparameter tuning, model selection, and early stopping procedures, thereby helping to mitigate overfitting to the training data.

Within the Ctrl-A framework, the validation dataset assumes an active role. First, it is used to measure the model response to the $K$ augmentation operations in the full augmentation pool $\mathcal{O}$, forming the basis for the update procedure for the ASD parameters $\boldsymbol{\Gamma}$ and $\boldsymbol{\alpha}$. Training data is deliberately excluded from the operation response analysis because the model is simultaneously optimized on augmented training samples. Instead, the validation dataset is used to obtain a more representative estimate of the effect of individual augmentation operations. Second, $\mathcal{D}_{\mathrm{Val}}$ enters the optimization objective in Eq.~\ref{eq:minimization}, where DA acts as an adaptive regularizer that balances the ratio between phase-aggregated training and validation losses toward a target setpoint, $\kappa_{sp}$. 

Despite its expanded role, the validation dataset is used strictly for guiding the control process and estimating operation responses. We observe no evidence that this usage induces overfitting to $\mathcal{D}_{\mathrm{Val}}$ as further analyzed in the experimental section and Appendix C.

\section{Experimental design and results}
\textbf{Framework and code.} Experiments are conducted using the PyTorch deep learning framework, with DA implemented through a customized version of the torchvision transformation library to enable the dynamic online augmentation required by Ctrl-A. The source code and implementation details are publicly available on GitHub \cite{controlaugmentrepo2026}.

\noindent\textbf{Experimental plan.}  Experiments are divided into two distinct phases: algorithm characterization (Sections~4.1-4.3) and comparative performance validation (Sections~4.4-4.6). In the initial phase, we use a lightweight benchmark architecture (airbench-94, \cite{jordan202494}) trained on CIFAR-10 to characterize the dynamic control mechanisms of the Ctrl-A algorithm. In addition, we investigate the algorithmic sensitivity to the number of operations applied per image ($N$) and the target control setpoint ($\kappa_{sp}$). In the validation phase, we evaluate the scalability and performance of Ctrl-A using deeper architectures, including Wide-ResNet-28-10 (WRN-28-10 \cite{zagoruyko2016wide}) on CIFAR-10/100 and SVHN-c, alongside ResNet architectures on Tiny ImageNet. Across the experiments, TA and RA serve as the baseline augmentation method, and all methods use identical training hyperparameters to ensure fair comparisons.

\noindent\textbf{Training procedure.} Models are trained using the cross-entropy loss function, and a stochastic gradient descent (SGD) optimizer with weight decay and Nesterov momentum. Training lasts $n_{max}$ epochs and uses a cosine learning-rate schedule of the form 
\begin{equation}
    \eta(n)=\frac{\eta_0}{2}\left(1 + ~\mathrm{cos} \left(\frac{\pi (n-1)}{n_{max}}\right)\right),
    \label{lr-schedule}
\end{equation}
in which $n$ is the epoch number and $\eta_0$ is the initial learning rate \cite{loshchilov2016sgdr}. Following completion of training, the final model is evaluated on the test set to obtain the per-run test accuracy. Unless otherwise stated, each experiment is independently repeated five times using different random seeds, and the reported results correspond to the mean with uncertainties given as the expanded uncertainty, computed as the standard error of the mean multiplied by the appropriate Student's t-factor for a $95\,\%$ confidence level.

\noindent\textbf{Model regularization.} To emphasize the effect of Ctrl-A, we exclude dropout functionality \cite{srivastava2014dropout}, and we avoid the use of $\ell_{1}$- and $\ell_{2}$-norm penalty terms in the loss function \cite{krogh1991simple,tibshirani1996regression}. This configuration is consistent with standard training practice for the considered architectures and ensures that data augmentation contributes substantially to the overall regularization, alongside implicit regularization from weight decay in the optimizer and architectural components such as batch normalization.

\noindent\textbf{Ctrl-A parameters.} We initialize $\Gamma_i^{(0)} = 0$ and $\alpha_i^{(0)} = 0$ for all $i$, set the training phase length  $n_p=5$, and $\Delta\gamma = 0.1$. The control parameter is initialized as $\xi^{(0)} = 0.9 $ irrespective of model size, dataset, and use of auxiliary transforms in the augmentation pipeline. The influence of this initialization is limited, as the closed-loop controller adjusts $\xi$ during the initial training phases given an appropriate choice of integral gain coefficient, $k_I$, as investigated in section 4.2. Additionally, we constrain the magnitude of the step size to the interval $|\Delta\xi | \in [0.005,0.1]$ to help ensure stable updates that are neither overly aggressive nor negligible, and enforce a lower bound $\xi_{\mathrm{min}} = 0.5$ to prevent excessively strong augmentations in cases where the control algorithm drives the system into saturation.

\noindent\textbf{Validation datasets.} A dedicated validation set is required for both model evaluation and method-specific configuration. For all experiments, we construct the validation set by holding out 1,000 samples from the original training sets using a fixed random seed shared across all runs, datasets, and methods. For RA, hyperparameters $(N,M)$ are selected based on this validation set following standard practice, whereas TA does not require validation-based tuning. In contrast, Ctrl-A uses the validation set exclusively to drive its closed-loop control mechanism. We empirically observe that a validation set of 1,000 samples provides sufficiently stable signals for both reliable ROR curve fitting and stable controller behavior, while remaining small relative to the full training set. The validation split is identical across all methods and is never used for test-time evaluation.

\subsection{Validation of control mechanism}
To validate the closed-loop behavior of Ctrl-A, we first conduct experiments using the airbench-94 architecture on CIFAR-10 \cite{jordan202494}. This experiment first focuses on characterizing the dynamics of the control system rather than comparing final classification performance.

\begin{figure}[ht]
    \centering
    \includegraphics[width=1\linewidth]{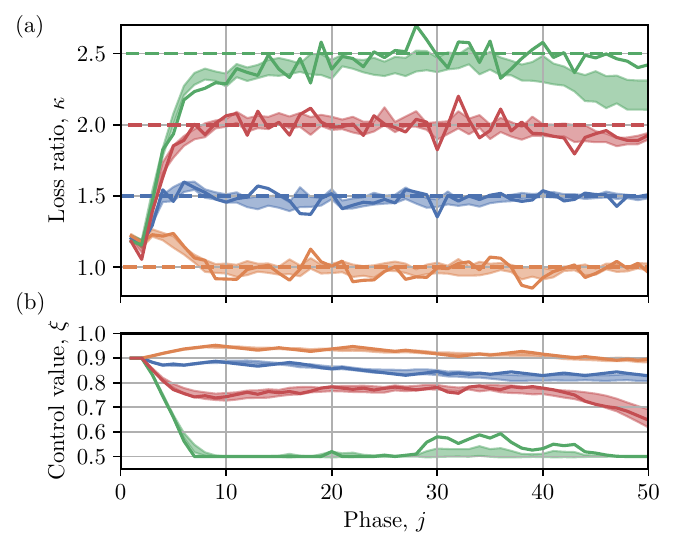}
    \caption{Evolution of the closed-loop control system using airbench-94. (a) Phase-averaged loss ratio $\kappa^{(j)}$ for different target setpoints $\kappa_{sp}$. Dashed lines denote the corresponding setpoints. (b) Evolution of the controller state $\xi^{(j)}$. The controller continuously adjusts $\xi$, and hence the ASD parameters, to regulate the loss ratio toward the desired operating point.}
    \label{fig:control}
\end{figure}

Figure \ref{fig:control} shows the evolution of the measured control signal $\kappa^{(j)}$ and the controller state $\xi^{(j)}$ across training phases for different values of the target setpoint $\kappa_{sp} \in \lbrace1.0, 1.5, 2.0, 2.5 \rbrace$. The solid curves show representative single-run trajectories, while the shaded regions indicate the ensemble mean together with the associated uncertainty estimated from five independent runs. This analysis allows us to assess (i) the ability of the controller to track the desired operating regime and (ii) the stability of the feedback loop. As expected, the control trajectories exhibit stochastic fluctuations. These fluctuations primarily arise from the stochastic nature of model optimization together with finite-sample variability in the phase-averaged validation signal. The proposed closed-loop controller successfully adjusts augmentation strengths to regulate the training dynamics. Furthermore, the controller converges toward distinct equilibrium values of $\xi$ for different target setpoints, demonstrating that Ctrl-A can consistently realize different augmentation regimes. An exception is the highest investigated setpoint, $\kappa_{sp}=2.5$, for which the controller reaches the lower bound $\xi=\xi_{\mathrm{min}}$ after only a few training phases. This indicates saturation of the controller, preventing further increases in augmentation strength.

\begin{figure}[ht]
    \centering
    \includegraphics[width=1\linewidth]{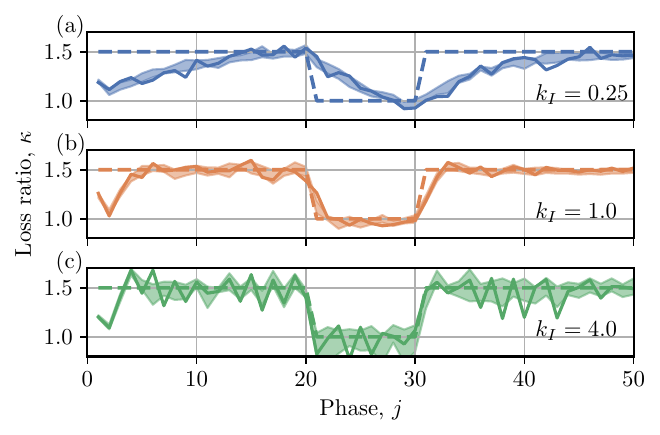}
    \caption{Influence of the integral gain coefficient, $k_I$, on the transient response of the Ctrl-A controller following step changes in the target setpoint, $\kappa_{sp}=1.5 \rightarrow 1.0 \rightarrow 1.5$. Dashed lines indicate the target setpoints. Solid curves show representative single-run trajectories of the measured feedback signal, $\kappa^{(j)}$, and shaded regions denote the ensemble mean $\pm$ standard error over seven independent runs.}
    \label{fig:gain}
\end{figure}

\subsection{Influence of the integral gain}
Having demonstrated that the controller can regulate the training dynamics toward different operating points, we next investigate the influence of the integral gain coefficient, $k_I$, on the transient response of the control system. In particular, we seek to characterize the trade-off between adaptation speed and control stability.

To this end, the target setpoint is varied during training according to a simple step sequence, $\kappa_{sp} = 1.5 \rightarrow 1.0 \rightarrow 1.5$, with transitions occuring after $40\,\%$ and $60\,\%$ of the total training duration. The resulting control trajectories of the measured loss ratio $\kappa^{(j)}$, are evaluated for three gain coefficients, $k_I \in \lbrace 0.25, 1.0, 4.0 \rbrace $. As shown in Fig. \ref{fig:gain}, the integral gain, $k_I$, governs the trade-off between adaptation speed and control stability. A low gain ($k_I=0.25$) results in slow convergence following setpoint changes, whereas a high gain ($k_I=4.0$) produces pronounced oscillations due to increased sensitivity to stochastic fluctuations. The intermediate value ($k_I=1.0$) provides a good balance between responsiveness and stability and is therefore adopted in the remaining experiments.

\subsection{Performance versus the setpoint, $\kappa_{sp}$}
Having established the control mechanism of Ctrl-A and determined an appropriate integral gain coefficient ($k_I = 1$), we next investigate how the target control setpoint, $\kappa_{sp}$, influences the final classification performance. In addition, we examine the interaction between $\kappa_{sp}$ and the number of augmentation operations, $N$, to identify suitable operating regimes.

\begin{figure}
    \centering
    \includegraphics[width=1\linewidth]{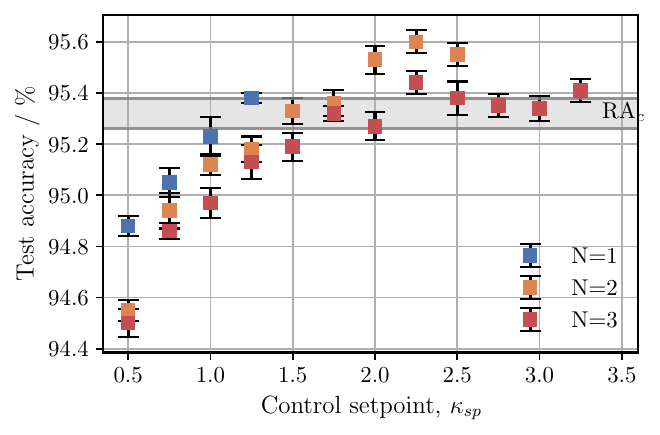}
    \caption{Top-1 test accuracy (model: airbench-94) on CIFAR-10 as a function of the target control setpoint $\kappa_{sp}$ and for $N= \lbrace1,2,3 \rbrace$ augmentation operations per image. Curves terminate when the controller reaches the lower bound $\xi_{\mathrm{min}}$, beyond which stronger augmentations cannot be realized. The shaded region indicates the ensemble mean $\pm$ standard error over 10 independent runs of the best performing benchmark method: search-optimized RandAugment using the Control augmentation space.}
    \label{fig:airbench94_vs_kappa_sp}
\end{figure}

Figure \ref{fig:airbench94_vs_kappa_sp} shows the resulting test accuracies for $N\in\lbrace 1,2,3 \rbrace$ as a function of the target control setpoint $\kappa_{sp}$. Test accuracy initially increases with $\kappa_{sp}$ for all three configurations, indicating that stronger augmentation-based regularization is beneficial. However, each curve eventually terminates as the controller saturates, i.e.~$\xi=\xi_{\mathrm{min}}$, beyond which stronger augmentations cannot be realized with the employed augmentation space from Appendix A. Evidently, the saturation point shifts towards larger values of $\kappa_{sp}$ as $N$ increases, reflecting that the overall augmentation strength is distributed across a greater number of operations. In general, the highest test accuracy is obtained for the smallest value of $N$ capable of realizing a given target setpoint. Nevertheless, among the investigated configurations, Ctrl-A($2$) achieves the highest overall test accuracy in the strong augmentation regime ($\kappa_{sp}\gtrsim2$), which could not be reached using a single augmentation operation.

Figure \ref{fig:airbench94_vs_kappa_sp} also includes the benchmark result obtained with search-optimized RA using our Control augmentation space of $(95.32\pm0.12)\,\%$. Additional benchmarks investigated (not shown) were search-optimized RA using the standard augmentation space $(94.96\pm0.14)\,\%$, TA with the Wide augmentation space $(94.96\pm0.12)\,\%$, and TA with the Control augmentation space $(95.21\pm0.17)\,\%$. The search-optimization for RA was performed on the two-dimensional grid $N\in\lbrace1,2,3,4\rbrace$, $M\in \lbrace7,12,17,22\rbrace$ for both augmentation spaces and using the constructed validation dataset for evaluation. Notably, a broad range of Ctrl-A configurations match or outperform all investigated baseline methods on this benchmark, indicating that competitive performance is achieved across a wide operating region rather than only at a narrowly tuned setpoint.


Based on these results, we adopt $N=2$ and $\kappa_{sp}=2$ as the default Ctrl-A configuration for the following benchmarking experiments. Importantly, this operating point is transferred unchanged across all benchmark evaluations, including different datasets and training setups, thereby assessing the robustness and generality of the proposed control strategy.

\subsection{Performance benchmarking on standard training setups}
\begin{table*}[tb]
     \caption{Top-1 test accuracy of WRN-28-10 on CIFAR-10, CIFAR-100, and SVHN-core. The first four columns are literature values \cite{zagoruyko2016wide,cubuk2019autoaugment,cubuk2020randaugment,muller2021trivialaugment}, whereas the last two columns correspond to the present work. Uncertainty values are reported with a coverage probability of $95\,\%$ and superscripts indicate the employed augmentation space: s: Standard \cite{cubuk2020randaugment}, w: Wide \cite{muller2021trivialaugment}, and c: proposed Control space (Appendix A). \label{tab:WRN2810standard}}
    \footnotesize
    \begin{center}
    \begin{tabular}{l|c c c c  | c c }
    & \multicolumn{4}{c| }{\textbf{Literature}}  &  \multicolumn{2}{c}{\textbf{This work}}   \\ 
    Dataset & Baseline & AA & RA$^{\mathrm{s}}$ & TA$^\mathrm{w}$  & TA$^\mathrm{c}$  & Ctrl-A$^{\mathrm{c}}$($N=2$)  \\  \hline
    CIFAR-10 & 96.1 & 97.4 & 97.3 & 97.46$\pm$.06 & 97.60$\pm$.07 & 97.53$\pm$.08  \\
    CIFAR-100 & 81.2 &  82.9 & 83.3  & 84.33$\pm$.17 &  84.03$\pm$.22 & 84.39$\pm$.23  \\
    SVHN-c & 96.9 & 98.1 & 98.3 & 98.11$\pm$.03  & 98.13$\pm$.05 & 98.33$\pm$.07 \\ \hline
    \end{tabular}
    \end{center}
    \raggedright
    \medskip 
\end{table*}
Having established a suitable operating regime for Ctrl-A, we next evaluate the method using the same parameter configuration on the standard benchmark protocols for CIFAR-10, CIFAR-100, and SVHN-core with the WRN-28-10 architecture (see Appendix D). Since these benchmark protocols are widely used for benchmarking DA methods, they provide a direct basis for assessing the performance of Ctrl-A against existing approaches. As the primary reference baseline, we employ search-free TA using the same (control) augmentation space. This allows the generality of the fixed Ctrl-A operating regime to be assessed against an established search-free augmentation strategy.

Table~\ref{tab:WRN2810standard} summarizes the benchmark results for the three datasets. The first four columns report representative literature results, including the baseline (minimal DA), AA, RA, and TA, whereas the final two columns present the corresponding results obtained in this work. We first note that TA employing the proposed Control augmentation space achieves performance comparable to the published TA benchmark based on the Wide augmentation space \cite{muller2021trivialaugment}, indicating that the proposed augmentation space itself is competitive. Using Ctrl-A with the fixed operating regime identified in the previous subsection ($N=2$, $\kappa_{sp}=2$), we achieve competitive performance that is comparable to or exceeds all included methods across the three benchmark protocols. The only exception is our CIFAR-10 result, where TA$^{\mathrm c}$ marginally outperforms Ctrl-A$^{\mathrm c}(2)$. The following subsection investigates this observation in greater detail.

\subsection{Analysis of the CIFAR-10 benchmark protocol}
\subsubsection{Effect of Cutout regularization}
The standard CIFAR-10 benchmark protocol includes an auxiliary 16-by-16 random Cutout transformation \cite{devries2017improved}, which constitutes a substantial source of augmentation-based regularization.  To investigate the influence of this component, we vary the Cutout magnitude while keeping all other training parameters fixed. Since Cutout is applied independently of the augmentation space, reducing its magnitude lowers the overall degree of augmentation-based regularization for a method like TA. 
\begin{figure}
    \centering
    \includegraphics[width=1\linewidth]{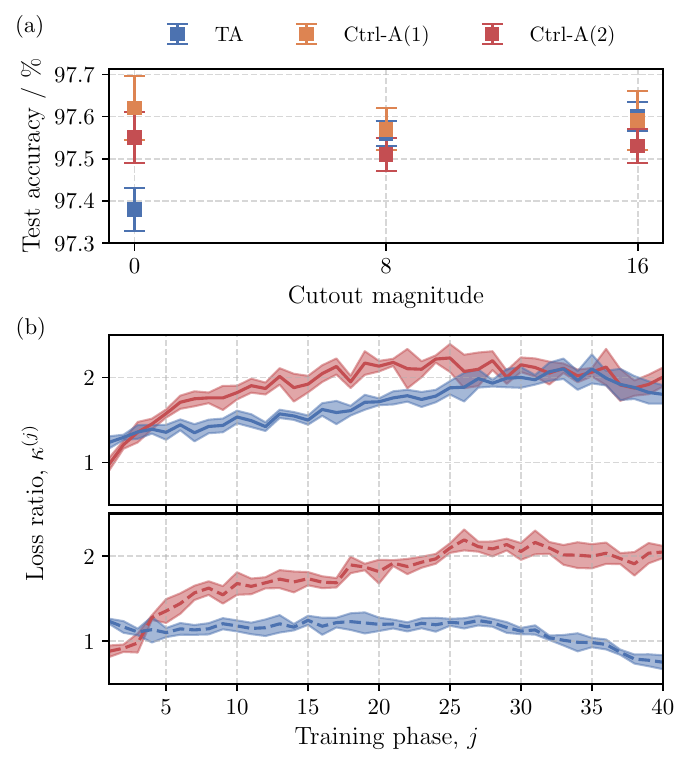}
    \caption{Efffect of Cutout magnitude. (a) Top-1 test accuracy of WRN-28-10 on CIFAR-10 using the standard benchmark protocol but with varying Cutout magnitude. (b) Measured loss-ratio trajectories, $\kappa^{(j)}$, for TA (blue) and Ctrl-A(2) (red) using Cutout magnitudes of 16 (top) and 0 (bottom). }
    \label{fig:cutout}
\end{figure}

Figure \ref{fig:cutout}(a) shows the test accuracy obtained for Cutout magnitudes ${0,8,16}$ using TA$^\mathrm{c}$, Ctrl-A$^\mathrm{c}$(1), and Ctrl-A$^\mathrm{c}$(2). The performance of TA is observed to decrease as the Cutout magnitude is reduced from the 16-by-16 transformation used in the standard benchmark configuration, whereas the performance of Ctrl-A remains largely unaffected. The corresponding loss-ratio trajectories, $\kappa^{(j)}$, in Fig.~\ref{fig:cutout}(b) explain this behavior. Without feedback, TA converges to a measured loss ratio of $\kappa^{(40)}\approx2$ when using the standard 16-by-16 Cutout (top plot), corresponding to operation in the strong augmentation regime. As the Cutout magnitude is reduced, the measured loss ratio shifts towards lower values, eventually resulting in slight overfitting when Cutout is removed entirely (bottom plot), reaching a loss ratio of $\kappa^{(40)}\approx0.75$. In contrast, Ctrl-A$(2)$ compensates for the reduced Cutout-induced regularization by increasing the augmentation strength parameters $\boldsymbol{\Gamma}$ and $\boldsymbol{\alpha}$, thereby maintaining an approximately constant operating point. 

For Ctrl-A$(1)$, the measured loss ratio likewise remains higher than that of TA as the Cutout magnitude is reduced, reaching $\kappa^{(40)}\approx1.0$ without Cutout. This behavior reflects the effect of the skewed ASDs (i.e.~$\alpha_i \neq 0 $), which shift the augmentation-strength distributions towards stronger perturbation magnitudes compared to the uniformly sampled augmentation strengths of TA.  Moreover, it is observed that Ctrl-A$(1)$ consistently achieves slightly higher accuracy than Ctrl-A$(2)$ as long as overfitting can be prevented by a single operation per image. This finding is consistent with the characterization study of Fig.~\ref{fig:airbench94_vs_kappa_sp}, where the highest performance for a given value of $\kappa_{sp}$ was generally obtained using the smallest number of operations, $N$, capable of realizing the desired augmentation-based regularization.

\subsubsection{Convergence with training duration}
The previous analysis established that the standard CIFAR-10 benchmark protocol operates in the strong augmentation regime. Under such conditions, the increased diversity of the augmented training data may permit continued model improvement over a longer training duration without inducing model overfitting. To test this hypothesis, we evaluate TA$^\mathrm{c}$, Ctrl-A$^\mathrm{c}(1)$, and Ctrl-A$^\mathrm{c}(2)$ using the standard benchmark protocol with the standard 16-by-16 Cutout transformation while varying the total number of training epochs, $n_{\max}$.
\begin{figure}
    \centering
    \includegraphics[width=1\linewidth]{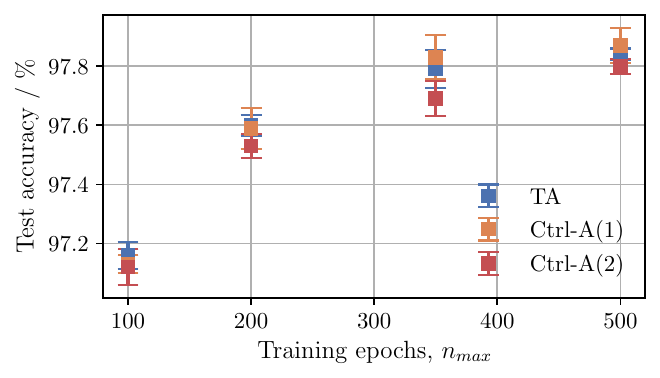}
    \caption{Top-1 test accuracy on CIFAR-10 as a function of the total number of training epochs, $n_{\max}$, for TA$^\mathrm{c}$, Ctrl-A$^\mathrm{c}(1)$, and Ctrl-A$^\mathrm{c}(2)$ using the standard benchmark protocol.}
    \label{fig:convergenceCIFAR10}
\end{figure}

Figure \ref{fig:convergenceCIFAR10} shows that all three methods continue to improve with increasing training duration throughout the investigated range. Statistically significant performance gains are observed from 200 to 350 epochs and marginal gains are observed from 350 to 500 epochs. This behavior is consistent with operation in the strong augmentation regime where the enhanced data diversity allows continued model improvement over an extended training duration without overfitting. The highest performance is observed for Ctrl-A$^\mathrm{c}(1)$ at a five run average of $97.87\,\%$ after training for 500 epochs. This suggests that the conventional CIFAR-10 benchmark protocol using 200 epochs does not fully exploit the potential of operation in the strong augmentation regime.

\subsubsection{Modified benchmark protocol}
The previous analysis in Sec.~4.5.1 demonstrated that the standard CIFAR-10 benchmark protocol derives a substantial fraction of its regularization from the auxiliary 16-by-16 Cutout transformation. While this configuration yields excellent performance, it also reduces the relative influence of the augmentation strategy itself, making comparisons between DA methods increasingly dependent on the particular choice of auxiliary regularization. Consequently, the performance of a DA method cannot be fully characterized by a single benchmark protocol. We therefore complement the standard benchmark with a modified protocol designed to reduce the influence of auxiliary regularization and more directly evaluate the augmentation strategy itself.

For the modified benchmark protocol, the training duration is increased from 200 to 500 epochs based on the analysis in Sec.~4.5.2, to better exploit the larger effective training diversity introduced by strong augmentations, while the auxiliary Cutout transformation is removed entirely. In addition, the initial learning rate and weight decay are reduced from 0.1 to 0.05 and from $5\times10^{-4}$ to $2.5\times10^{-4}$, respectively, reflecting a reduced reliance on learning-parameter based regularization. Finally, nearest-neighbor pixel interpolation is replaced by bilinear interpolation for geometric transformations to improve image fidelity under strong augmentations.


\begin{table}[h]
     \caption{Top-1 test accuracy of WRN-28-10 on CIFAR-10 using the modified benchmark protocol and different augmentation spaces. \label{tab:WRN2810modified}}
    \footnotesize
    \begin{center}
    \begin{tabular}{l |c c c  }  
    Aug.~space & TA & Ctrl-A$(1)$ & Ctrl-A($2$)  \\  \hline
    Standard & 97.51$\pm$0.10 & 97.60$\pm$0.12   & 97.76$\pm$0.14 \\
    Wide & 97.81$\pm$0.09  & 97.96$\pm$0.12 & 98.01$\pm$0.10   \\ 
     Control & 97.89$\pm$0.18  & 98.01$\pm$0.06 & 98.20$\pm$0.09  \\\hline
    \end{tabular}
    \end{center}
\end{table}

Table \ref{tab:WRN2810modified} shows the results obtained with the proposed modified benchmark protocol comparing TA and Ctrl-A across the Standard, Wide, and Control augmentation spaces to investigate both the effect of the benchmark protocol and the contribution of the augmentation space. 

The Standard augmentation space produces the lowest performance across all three augmentation configurations. This can be explained by the significant overfitting observed under the modified protocol using this augmentation space. Even for Ctrl-A(2) the ASD parameters become saturated in the early training phases due to the limited augmentation strengths available within this space.

For the Wide augmentation space, none of the methods exhibit overfitting, however, Ctrl-A still outperforms TA for both $N=1$ and $N=2$. In particular, for Ctrl-A, we observe that the ASD parameter $\Gamma$ remains below unity throughout training for select operations, namely translations, rotations and solarization. This likely contributes to Ctrl-A outperforming TA for both $N=1$ and $N=2$. 

The best overall result is obtained with the Control augmentation space, reaching $98.20\,\%$ for Ctrl-A($2$). This represents a significant improvement over the standard benchmark protocol (see Table~\ref{tab:WRN2810standard} and Fig.~\ref{fig:convergenceCIFAR10}), whereas the performance of TA remains at a similar level. Overall, these results demonstrate the benefit of adaptive augmentation but they also establish the effectiveness of the proposed Control augmentation space.

\begin{figure}
    \centering
    \includegraphics[width=1\linewidth]{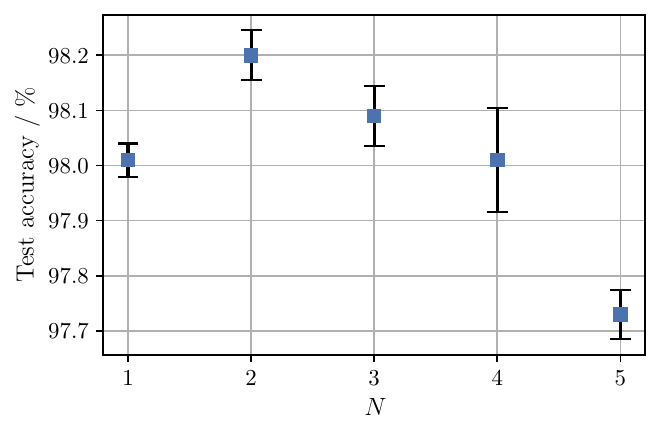}
    \caption{Top-1 test accuracy of WRN-28-10 on CIFAR-10 as a function of the number of operations, $N$, for Ctrl-A using the modified benchmark protocol and the Control augmentation space. \label{fig:mod_vs_N}}
\end{figure}
To finally investigate the effect of the operation number, $N$, we evaluate the performance of Ctrl-A for $N=\lbrace 1,2,3,4,5 \rbrace$ using the Control augmentation space and the modified benchmark training protocol. The results of this investigation are shown in Fig.~\ref{fig:mod_vs_N}, illustrating best performance, in this case, is obtained for $N=2$. Despite increasing the available control authority, increasing $N$ beyond this value degrades performance slightly for $N=3$ and $N=4$ before decreasing substantially for $N=5$. These results suggest that increasing the number of controllable operations does not necessarily translate into improved generalization performance.

\subsection{Tiny ImageNet generalization}
Our final experiments evaluate Ctrl-A as we move from the low resolution ($32\times32$) image vision tasks of CIFAR and SVHN-c to the higher resolution ($64\times 64$) Tiny ImageNet dataset \cite{le2015tiny}. We here employ both ResNet-18 and ResNet-50 architectures while retaining the Ctrl-A parameters $N=2$ and $\kappa_{sp}=2.0$, thereby further assessing the transferability of these parameters across datasets and classification model types. 

For the training pipeline, individual models are optimized from scratch, i.e.~without ImageNet pre-training, over 200 epochs using a standard Tiny ImageNet training configuration with an initial learning rate of $\eta_0 = 0.1$ and a weight decay of $5\times10^{-4}$. To adjust for the 64-by-64 input resolution, we replace the standard ResNet stem consisting of a $7\times7$, stride-2 convolution and subsequent max-pooling layer with a $3\times3$, stride-1 convolution, following common adaptations of ResNet for Tiny ImageNet \cite{residual2016tinyimagenet,zeyuanyin_tiny_imagenet}. For auxiliary DA, we incorporate random horizontal flipping, 8-pixel pad-and-crop, and moderately aggressive 24-by-24 Cutout to mitigate overfitting given the limited number of samples per class in the Tiny ImageNet training set. 

Our comparative results on Tiny ImageNet for Ctrl-A, benchmarked against TA using the Control augmentation space, are shown in Table~\ref{tab:TinyImagenet}. Despite being trained from scratch, both ResNet architectures exceed $70\,\%$ top-1 accuracy using Ctrl-A(2). In both cases, Ctrl-A(2) outperforms the corresponding TA baseline with improvements of 0.81 and 0.78 percentage points for ResNet-18 and -50, respectively. Notably, the ResNet-50 experiment with Ctrl-A(2) exhibited comparatively higher run-to-run variations, which was primarily caused by a single run yielding a significantly lower test accuracy than the remaining four runs. However, the apparent improvement relative to TA across both architectures indicates that the Ctrl-A framework is not limited to low resolution image classification tasks such as CIFAR and SVHN.   

\begin{table}[h]
     \caption{Top-1 test accuracy of ResNet-18 and ResNet-50 on Tiny Imagenet (no pre-training). \label{tab:TinyImagenet}}
    \footnotesize
    \begin{center}
    \begin{tabular}{l |c c   }  
    Model & TA & Ctrl-A($2$)  \\  \hline
    ResNet-18 & 69.74$\pm$0.24   & 70.55$\pm$0.29 \\
    ResNet-50 & 71.36$\pm$0.25 & 72.14$\pm$0.75   \\ \hline
    \end{tabular}
    \end{center}
\end{table}

\section{Discussion}
\subsection{Ctrl-A as online adaptive regularization}
The results obtained throughout the previous section supports our formulation of DA as an online adaptive regularization mechanism. By adjusting augmentation strengths during training based on the dynamic relationship between training and validation performance, Ctrl-A allows the degree of DA-based regularization to change during the learning process. This provides an alternative to determining a fixed augmentation policy before training and suggests that adaptation to the model's evolving state can be a useful principle for automated DA.

\subsection{Computational efficiency}
Both offline- and online adaptive DA methods are often bottlenecked by substantial computational costs \cite{cubuk2019autoaugment,lim2019fast,ho2019population,lin2019online,tang2020onlineaugment}. In contrast, Ctrl-A introduces very limited overhead. Empirically, a single-run training execution with Ctrl-A incurs a computational overhead of approximately $10\,\%$ compared to the static, tuning-free TrivialAugment (TA) benchmark \cite{muller2021trivialaugment}. The computational cost of the update procedure scales linearly with the size of the augmentation space, $K$, the validation dataset size, and the relative frequency of the update procedure, $n_{\mathrm{max}} / n_p$, while scaling inversely with the step size, $\Delta \gamma$, used to construct the ROR curves. Crucially, because Ctrl-A calculates these curves directly via rapid forward passes on a relatively small held-out validation subset rather than full-scale training loops, the additional computational cost remains modest in practice. This overhead can be reduced even further by caching the ROR curve re-computation, e.g.~limiting it to specific macro-phases of the control mechanism, while updating only the core scalar control parameter, $\xi $, and the ASD parameters at every individual phase.

\subsection{Implications for benchmarking automated DA}
Our convergence study, performed for CIFAR-10, revealed that the standard WRN-28-10 training pipeline commonly used to benchmark DA methods can be sub-optimal when exploiting strong DA. Evidently, the standard 200-epoch pipeline provides too few training iterations to reach convergence when strong DA substantially increases training-data variability. As noted by Hounie et al.~\cite{hounie2023automatic}, the training setup used for benchmarking should not itself become a performance bottleneck when comparing DA methods. In this case, the standard WRN-28-10 training protocol for CIFAR-10 was originally developed for a setting with relatively limited augmentation, and may therefore be less well suited to evaluating strong DA strategies.

With our modified and prolonged training setup, we demonstrate that WRN-28-10 achieves CIFAR-10 performance comparable to results obtained using the substantially more computationally expensive Shake-Shake-model training pipeline \cite{muller2021trivialaugment}. The impact of the training setup is particularly evident for Ctrl-A with $N=2$, for which changing from the standard to the modified benchmark protocol reduced the CIFAR-10 error rate by approximately 25$\,\%$. However, the same modified protocol causes severe overfitting when the strong DA and Cutout components are removed, despite retaining conventional random cropping and horizontal mirroring, reducing the final test accuracy to approximately \mbox{92--93\,$\%$}. Our results therefore highlight that training configurations, including choices of regularization hyperparameters such as weight decay and auxiliary data transformations, can substantially affect the relative performance of DA methods when the training setup itself becomes the performance bottleneck.

\subsection{Practical considerations}
An important practical property of Ctrl-A is that the control objective is expressed in terms of the desired regularization behavior rather than directly in terms of augmentation magnitudes. Across the benchmark experiments, we use the same control setpoint, $\kappa_{sp} =2$, while Ctrl-A adapts the underlying augmentation-strength distributions to the response of the model. This differs from RA \cite{cubuk2020randaugment}, for which the optimal augmentation magnitude $M$ typically depends on both the classification task and experimental setting. However, a fixed augmentation-strength setup generally does not provide the same effective degree of regularization across datasets, models, or training configurations. Our results therefore suggest that parametrizing DA in terms of a desired regularization level, rather than fixed augmentation strengths, may provide a more robust basis for configuring DA across different classification tasks and training configurations.

With respect to Ctrl-A hyperparameters, our experiments suggest that increasing the number of operations, $N$, does not necessarily improve performance, even when it provides greater control authority and allows $\kappa$ to track $\kappa_{sp}$ more closely. In practice, the best performance was generally obtained with a low number of operations, with $N=2$ generally providing a good compromise between the number of operations and control authority. However, the $N=1$ case was also observed to perform well, even in cases where $\kappa$ could not remain close to $\kappa_{sp}=2.0$, and showed particularly strong performance in cases with strong auxiliary transformations such as Cutout. Furthermore, while we used $\kappa_{sp}=2.0$, throughout the benchmark experiments, our results do not establish that this value is generally optimal across datasets and models. Nevertheless, we consistently observe better generalization performance when training in the strong-augmentation regime ($\kappa_{sp}>1$) compared to training in the weak-augmentation regime ($\kappa_{sp}<1$).


\subsection{Limitations and future directions}
The Ctrl-A algorithm can be interpreted as a feedback control system in which the strength of data augmentation serves as the control input that is used to regulate the ratio between training and validation losses \cite{aastrom2021feedback}. In this view, the ASD parameters $\boldsymbol{\Gamma}$ and $\boldsymbol{\alpha}$ act as the control loop actuators, and, as in any feedback system, actuator saturation may occur. In our framework, saturation occurs if $\xi \rightarrow \xi_{\mathrm{min}}$, causing the ASD parameters to reach their limiting values and thereby bounding the available control authority of the algorithm. Consequently, the process variable, $\kappa$, can in this case fail to reach its setpoint value, $\kappa_{sp}$, which was observed in a subset of the experiments, particularly those using the Standard augmentation space on CIFAR-10 and in the CIFAR-100 experiments.

To counter cases like this, we envisage incorporating an extra control layer into the algorithm, either by dynamically increasing the number of operations, $N$, or by adapting the underlying augmentation pool to one that provides greater control authority. Our experiments indicate that the Wide augmentation pool provides the greatest control authority, followed by the proposed Control pool, while the Standard pool provides the least.

In our experiments we generally observe the tendency that increasing the number of operations $N$ beyond 2 leads to either performance stagnation or degradation. Although the reason for this is not yet fully understood, we hypothesize that interactions between combinations of transformations, such as translations, shears, and rotations, may lead to augmented samples that retain only a limited portion of the original image information. Such interactions could be investigated by extending the ROR analysis to quantify model responses to pairs of augmentation operations. Accounting for such interactions could allow Ctrl-A to suppress combinations of transformations that are detrimental to model training.

Another path for extending Ctrl-A would be to account for class-dependent model sensitivities to the $K$ operations in the augmentation space. Currently, the augmentation strength of each operation is updated independently of the class labels of the augmented images. However, different classes in a given image classification task may exhibit different sensitivities to image transformations. A well known example is that of digit classification, in which certain digits are more sensitive to  rotations than others \cite{mo2024ric}. This limitation may be particularly relevant for imbalanced datasets, for which dataset-wide performance measures can mask augmentation effects that are detrimental to minority classes. Therefore, a natural extension of Ctrl-A is to reformulate $\boldsymbol{\Gamma}$ and $\boldsymbol{\alpha}$ as matrices of size $K\times C$ rather than 1-dimensional arrays of length $K$. 

All experiments reported in this work were conducted on a single-GPU workstation equipped with an NVIDIA RTX 4070 Ti SUPER 16 GB. The computational resources available in this setup precluded evaluation of Ctrl-A on large-scale datasets such as ImageNet-1k and limited benchmarking of modified training configurations to the TA method (see Tables~\ref{tab:WRN2810standard} and \ref{tab:WRN2810modified}). Access to additional computational resources would enable evaluation of Ctrl-A in larger-scale and more demanding settings in terms of parallel processing power and memory.\\

\subsection{Concluding remarks}
We have introduced ControlAugment (Ctrl-A), a control-driven algorithm that realizes online model regularization through adaptive data augmentation. To achieve this, we expand the purpose of the conventional validation dataset to encompass computations of what we define as relative operation response curves. These response curves form the basis for a control-based update procedure for each individual operation type in the chosen augmentation space. With Ctrl-A, we demonstrate highly competitive performance in standard benchmarking tasks that use the Wide-ResNet model architecture. Notably, this is achieved without resource overhead from a separate search phase and while simply initializing all augmenting transformations to the identity operator. 

\section{Acknowledgement}
This work was supported by; Project 22HLT05 MAIBAI, which has received funding from the European Partnership on Metrology, co-financed from the European Union's Horizon Europe Research and Innovation Program and by the Participating States; by Innovate UK under the Horizon Europe Guarantee Extension; and by the Danish Agency for Higher Education and Science. 

\bibliographystyle{unsrt}
\bibliography{biblo} 

\clearpage

\appendix

\appendixpage

\section{Control augmentation space}
Table \ref{tab:IAaugpool} contains the $K=15$ transformations in our ``Control'' augmentation space. The first six, \textsc{TranslateX}, \textsc{TranslateY}, \textsc{ShearX}, \textsc{ShearY}, \textsc{Scale}, and \textsc{Rotation} fall into the category of geometric (affine) transformations, whereas the last nine, \textsc{Hue}, \textsc{Bright/dark}, \textsc{Sharpen/blur}, \textsc{Contrast}, \textsc{Saturation}, \textsc{Solarize}, \textsc{Posterize}, \textsc{AutoContrast}, and \textsc{Equalize} are appearance-based (color) transformations.

\begin{table}[ht]
    \centering
    \caption{Operations in the ``control" augmentation space and respective parametrizations that use normalized augmentation strength $\gamma_i \in [0,1]$. All operations converge to the identity operator for $\gamma  \rightarrow 0$, and the identity operator is therefore explicitly excluded from the augmentation space. Operations appended by (s) indicates that the sign of $\gamma_i $ is flipped, i.e.~$\gamma_i < 0$, with a $50\,\%$ probability.}
    \label{tab:IAaugpool}
    \begin{tabular}{c|c|c}
    \Xhline{1.5pt}
     Op.~$\#$  & Op.~Name & Parametrization 
      \\ \Xhline{1.5pt}  
      1. & TranslateX (s) & $\gamma_1/2\times$Img.\,width
      \\
      2. & TranslateY (s) & $\gamma_2/2\times$Img.\,height 
      \\
      3. & ShearX (s) &  $45^\circ\gamma_3$ 
      \\
      4. & ShearY (s) &  $45^\circ\gamma_4$ 
      \\
      5. & Scale (s) & $1 + \gamma_5/2$ 
      \\
      6. & Rotation (s) & $60^\circ\gamma_6 $
      \\
      7. & Hue (s) &  $\gamma_{7}/2$
      \\ 
      8. & Bright/dark (s) &  $1+ 0.9\gamma_8$
      \\
      9. & Sharpen/blur (s) &  $1 +0.9 \gamma_{9} $
      \\
      10. & Contrast (s) &  $1 + 0.9\gamma_{10} $
      \\
      11. & Saturation (s)  & $1 + 0.9\gamma_{11}$
      \\
      12. & Solarize & $255(1-\gamma_{12}/2)$  
      \\
      13. & Posterize & $8(1- \gamma_{13}/2 )$
      \\
      14. & AutoContrast & $\gamma_{14}$
      \\
      15. & Equalize & $\gamma_{15}$
      \\
      \Xhline{1.5pt}
    \end{tabular}
\end{table}
The parametrization of each operation (based on transformations from the torchvision.transforms python library), is provided in the right column of the table. As apparent from the parametrization of solarize and posterize, we assume 8-bit images. In addition most operation types, i.e.~those appended with (s), are signed operations, meaning that $\gamma_{1\textrm{-}11} \in [-1,1]$. The sign of $\gamma_{1\textrm{-}11}$ is determined by a probabilistic flip of probability 50\,$\%$. 

AutoContrast ($\mathcal{O}_{14}$) and Equalize ($\mathcal{O}_{15}$) are usually not treated as parametrized operations \cite{cubuk2019autoaugment,cubuk2020randaugment}, and so our implementation requires a short explanation. To construct parametrized versions, we use linear combinations
\begin{equation}
    \mathcal{O}_{14,15}(x;\gamma) = (1-\gamma) x + \gamma  \,\mathcal{O}_{14,15}'(x),
\end{equation}
in which $\mathcal{O}'_{14}$ is the standard non-parametrized version of AutoContrast and $\mathcal{O}'_{15}$ is the standard non-parametrized version of Equalize. In this form, operations $\mathcal{O}_{14}$ and $\mathcal{O}_{15}$ naturally adhere to the general rules for operations in ControlAugment.



\begin{figure}
    \centering
    \includegraphics[width=0.98\linewidth]{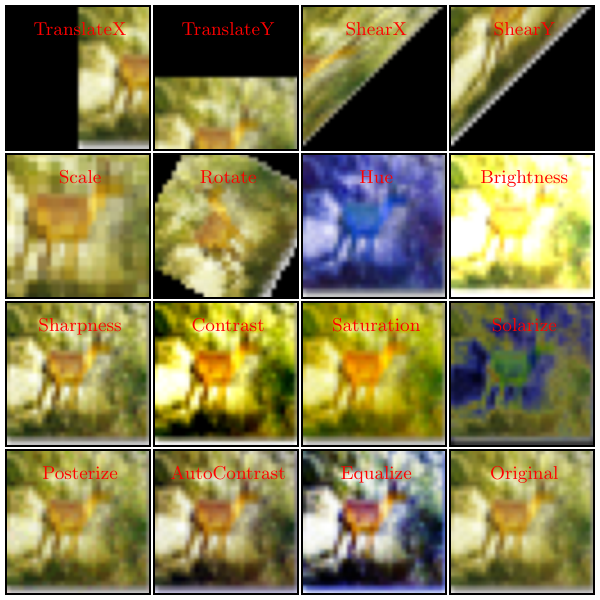}
    \caption{Visualization of uni-transformed images for each of the operations in the full augmentation space with augmentation strength $\gamma=+1$. The original image is shown in the lower right.  }
    \label{fig:max_aug_example}
\end{figure}

Figure \ref{fig:max_aug_example} illustrates uni-transformed versions ($\gamma=1$) of a single CIFAR-10 image for each of the 15 operations in the Control augmentation space. These examples are not necessarily representable for augmented images, but they showcase the maximum perturbation an image can undergo for each operation $\mathcal{O}_i $ if $\gamma_i = 1$. 

\section{Augmentation strength updates}
This section details the augmentation strength update algorithm used to solve Eq.~\ref{eq:IArule} which is repeated here for convenience:
\begin{equation}
    \mathrm{Acc}_f (\mathcal{O}_i (\mathcal{D}_\mathrm{Val}  ; \gamma )) / \mathrm{Acc}_f (\mathcal{D}_\mathrm{Val} ) - \xi = 0,
\end{equation}
where the first term is defined as the relative operation response (ROR). 
The evaluation uses different degrees of augmented versions of the validation data to determine, for each operator type $\mathcal{O}_i$, the augmentation strength that leads to a decrease in accuracy of $\xi \in [0,1]$. Our implementation uses a small fraction (1000 samples, corresponding to $10~\%$ for CIFAR-10) of either the test dataset or the training dataset. 

\begin{figure}
    \centering
    \includegraphics[width=0.75\linewidth]{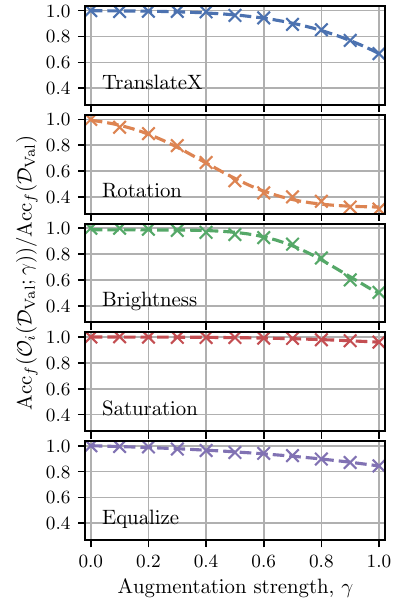}
    \caption{Examples of relative operation response data (crosses) and fitting curves (dashed lines) used to determine $\Gamma_i$ and $\alpha_i$. CIFAR-10 was used as example dataset.}
    \label{fig:appendixB}
\end{figure}

Our algorithm works as follows. For each operation $\mathcal{O}_i$, prepare the datasets $\mathcal{O}_i(\mathcal{D}_\mathrm{Val}; \gamma)$ for $\gamma=\{0.1 , 0.2, \dots 0.9, 1\}$ and compute, for each operation and each $\gamma $, the model accuracy $\mathrm{Acc}_f (\mathcal{O}_i (\mathcal{D}_\mathrm{Val}  ; \gamma ))$. For each operation, this results in data arrays in which the augmentation strength $\gamma$ is the independent variable, providing us with a basis for (approximately) solving the implicit equation above. To accomplish this, we make use of regression analysis as exemplified in Fig.~\ref{fig:appendixB} for different types of operations. The error function provides an adequate, and natural, model fit to describe an increasing error due to an increasing amount of perturbation. 


\subsection{Linearly tilted distribution}
The original implementation of Ctrl-A simply drew augmentation strengths from a uniform distribution according to $\gamma_i \sim U(0, \Gamma_i )$. However, it was quickly discovered, that this simple construction provided insufficient control capacity for larger models, such as the WRN-28-10 model variant, which ended up resulting in $\Gamma_i = 1 $ for all $i$ and the model overfitting to the training data in the fine tuning phases. To mitigate this, we introduce the possibility of a tilt $\alpha_i$ to the distributions for which $\Gamma_i = 1$, such that the augmentation strengths in these cases are instead drawn from ``tilted'' distributions, i.e.~$\gamma_i \sim U_{\alpha_i} (0,1)$, which are exemplified in Fig.~\ref{fig:ASDs} in the main text. This modification enables more control authority by placing more weight on stronger augmentation strengths for those transformation types that are efficiently learned by the model. 

In cases where the model $f_\theta$ is $\xi$-insensitive to operation $\mathcal{O}_i$ such that $\Gamma_i = 1$, we compute the corresponding tilt parameter $\alpha_i$ as
\begin{equation}
    \alpha_i = \frac{\mathrm{Acc}_f (\mathcal{O}_i (\mathcal{D}_\mathrm{Val}  ; \gamma = 1 )) /\mathrm{Acc}_f (\mathcal{D}_\mathrm{Val} )-\xi}{1-\xi} .
\end{equation}
With this construction, $\alpha_i = 0$ if Eq.~\ref{eq:IArule} is exactly solved for $\gamma = 1$, and $\alpha_i = 1$ if the transformation type does not degrade model performance at all such that  $\mathrm{Acc}_f (\mathcal{O}_i (\mathcal{D}_\mathrm{Val}  ; \gamma = 1 )) = \mathrm{Acc}_f (\mathcal{D}_\mathrm{Val} ) $.



\section{Testing for overfitting}
\begin{figure}[hb!]
    \centering
    \includegraphics[width=1\linewidth]{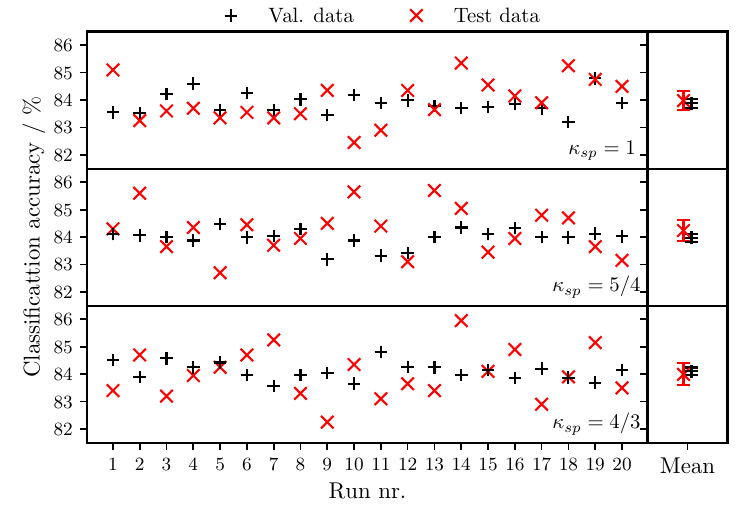}
    \caption{Comparison study exploiting an 8000/2000 validation-test split to investigate whether the Ctrl-A algorithm leads to overfitting to the validation data used to update augmentation strengths. The results are obtained by training a (custom) LeNet model on CIFAR-10 employing Ctrl-A(2) augmentation.}
    \label{fig:nooverfitting}
\end{figure}

Due to the Ctrl-A procedure invoking online feedforward based on augmented validation data to tune the available augmentation strength distributions, it should be ensured that the process does not lead to undesired overfitting of the model to the validation data. To this end, we create (randomized) 8000/2000 validation/test splits of the 10k CIFAR-10 test samples, and use the larger pool of 8000 validation images to inform the data augmentation. Our results, obtained for a 568-parameter LeNet model trained on CIFAR-10 using the CtrlA(2) policy, are visualized in Fig.~\ref{fig:nooverfitting} for three different values of $\kappa_{sp}$. 

To statistically assess the data in Fig.~\ref{fig:nooverfitting} we use a one-sided Welch's $t$-test with the null hypothesis, $H_0:$  $\mathbb{E}[\mathrm{Acc}_f(\mathcal{D_\mathrm{Val}})] \leq \mathbb{E}[ \mathrm{Acc}_f(\mathcal{D_\mathrm{Test}}) ]$
and the alternative hypothesis $H_1:$  $\mathbb{E}[\mathrm{Acc}_f(\mathcal{D_\mathrm{Val}})] > \mathbb{E}[ \mathrm{Acc}_f(\mathcal{D_\mathrm{Test}}) ]$. The observed $p$-values ($0.69$, $0.89$, and $0.31$) provide no significant evidence against $H_0$ which cannot be rejected at any conventional significance level. Thus, we find no statistical support for the validation accuracy being higher than the test accuracy, consistent with our expectation that the ControlAugment algorithm does not lead to the model overfitting to the validation data.


{
\begin{table*}[!t]
    \caption{Setup parameters and auxiliary transforms used for the experiments with CIFAR, SVHN-c, and Tiny ImageNet. Random crop, RC(N): N-pixel random pad-and-crop, H-flip: random horizontal flips, flips append: horizontal flips appended to training set, Norm.: Image normalization, Cutout(N): N-by-N random cutout.  }
    \label{tab:CIFARsetup}
    \centering
    \small
    \begin{tabular}{l| c|c|c| c| c }
    Dataset  & Epochs & LR & WD & Pre-transforms & Post-transforms   \\\hline
    CIFAR & 200 & 0.1 & $5\times10^{-4}$  &  RC(4) + H-flip  & Norm. + Cutout(16)  \\
   
   SVHN-c & 300 & 0.005 & 0.005  & - &  Norm. + Cutout(16) \\ 
   Tiny ImageNet & 200 & 0.1 & $5\times10^{-4}$ & RC(8) + H-flip &  Norm. + Cutout(24) \\ 
   Modified, CIFAR-10 & 500 & 0.05 & $2.5\times10^{-4}$ & RC(4) + flips append & Norm. \\
   \hline
    \end{tabular}
    \vspace{0.5cm}
\end{table*}
}

\section{Dataset details}
\label{appD}

\subsection{CIFAR} 
CIFAR (Canadian Institute For Advanced Research) are benchmark image classification datasets consisting of downsampled (32-by-32) real-world RGB images containing centered objects (e.g. airplane, cat, ship, truck, etc.). The CIFAR dataset comes in two variants; the 10-class dataset CIFAR-10, and the 100-class dataset CIFAR-100. Each of the two datasets contain 50,000 training samples and 10,000 test samples. 

In our training setups for CIFAR, WideResNet-28-10 models are trained using stochastic gradient descent with a Nesterov momentum of 0.9 and the cosine learning-rate schedule in Eq.~\ref{lr-schedule}. The initial learning rates (LR), weight decay (WD), auxiliary pre-transformations and post-transformations are listed in 
Table~\ref{tab:CIFARsetup} for the standard and modified setups. Pre-transformations are those operations that are applied before operations from the chosen augmentation space, and post-transformations are those that are applied after.

\subsection{SVHN-core} 
SVHN-core (Street View House Numbers - Core) is a real-world image dataset containing 32-by-32 pixel images of house number digits (0-9). The dataset contains 73,257 training samples and 26,032 test samples, and exhibits strong class imbalance as certain digits, especially ``1", naturally appears more frequently than others in house numbers.

In our training setups for SVHN-c, WideResNet-28-10 models are trained using stochastic gradient descent with a Nesterov momentum of 0.9 and the cosine learning-rate schedule in Eq.~\ref{lr-schedule}. The initial learning rates (LR), weight decay (WD), auxiliary pre-transformations and post-transformations are listed in 
Table~\ref{tab:CIFARsetup}.



\subsection{Tiny ImageNet} 
The Tiny ImageNet dataset is a downsampled ($64\times64$) subset of the full ImageNet 1k dataset. It contains 100,000 labelled training images,  10,000 labelled validation images, and 10,000 unlabelled test images. As is common practice when using Tiny ImageNet for supervised learning, we exclude the test images and instead use the validation set as the practical test set. The labelled training and validation sets contain 200 different object classes and are both class-balanced.

In our training setups for Tiny Imagenet, stem-modified ResNet-18 and ResNet-50 models are trained from scratch using stochastic gradient descent with a Nesterov momentum of 0.9 and the cosine learning-rate schedule in Eq.~\ref{lr-schedule}. The initial learning rates (LR), weight decay (WD), auxiliary pre-transformations and post-transformations are listed in Table~\ref{tab:CIFARsetup} for the standard and modified setups.



\end{document}